%% file: root.tex
\documentclass[letterpaper, 10 pt, conference]{ieeeconf}  % Comment this line out if you need a4paper

\IEEEoverridecommandlockouts                              % This command is only needed if 
\usepackage{amsmath} % assumes amsmath package installed
\usepackage{siunitx}
\usepackage{algorithm}
\usepackage{amsfonts}
\usepackage{verbatim}
\usepackage[noend]{algpseudocode}
\usepackage[none]{hyphenat}
\usepackage{graphicx}
\usepackage{pgf}
\usepackage{pgfplots}
\usepackage{subfig}
\usepackage{cite}
\usepackage{hyperref}
\usepackage{balance}
\usepackage{textpos}

\usepackage{mathtools}
\let\labelindent\relax
\usepackage[inline]{enumitem}

\graphicspath{{images/}}

\DeclarePairedDelimiter\abs{\lvert}{\rvert}%
\DeclarePairedDelimiter\norm{\lVert}{\rVert}%

\usepgfplotslibrary{external}
\algnewcommand\algorithmicforeach{\textbf{for each}}
\algdef{S}[FOR]{ForEach}[1]{\algorithmicforeach\ #1\ \algorithmicdo}
\DeclareMathOperator*{\argmin}{arg\,min}

\title{\LARGE \bf
	Joint Point Cloud and Image Based Localization \\ For Efficient Inspection in Mixed Reality
}

\author{Manash Pratim Das$^{1}$, Zhen Dong$^{2}$ and Sebastian Scherer$^{3}$% <-this % stops a space
	\thanks{$^{1}$Manash Pratim Das is with the Indian Institute of Technology, Kharagpur, 721302, WB, India. Email: {\tt\small mpdmanash@iitkgp.ac.in}}%
	\thanks{$^{2}$Zhen Dong is with the State Key Laboratory of Information Engineering in Surveying, Mapping and Remote Sensing, Wuhan University, Wuhan 430079, China. Email: {\tt\small dongzhenwhu@whu.edu.cn}}%
	\thanks{$^{3}$Sebastian Scherer is with the The Robotics Institute, Carnegie Mellon University, 5000 Forbes Ave, Pittsburgh, PA 15213, USA. Email: {\tt\small basti@andrew.cmu.edu}}%
}

\begin{document}
	\maketitle
	\thispagestyle{empty}
	\pagestyle{empty}
	\begin{textblock*}{\textwidth}(0cm,-5cm)
		\centering
		\textcolor{gray}{This paper has been accepted for publication at the IEEE/RSJ International Conference on\\
		Intelligent Robots and Systems (IROS), Madrid, 2018. \copyright IEEE}
	\end{textblock*}
	\vspace{-1.6em}
	%%%%%%%%%%%%%%%%%%%%%%%%%%%%%%%%%%%%%%%%%%%%%%%%%%%%%%%%%%%%%%%%%%%%%%%%%%%%%%%%
	\begin{abstract}
		This paper introduces a method of structure inspection using mixed-reality headsets to reduce the human effort in reporting accurate inspection information such as fault locations in 3D coordinates. Prior to every inspection, the headset needs to be localized. While external pose estimation and fiducial marker based localization would require setup, maintenance, and manual calibration; marker-free self-localization can be achieved using the onboard depth sensor and camera. However, due to limited depth sensor range of portable mixed-reality headsets like Microsoft HoloLens, localization based on simple point cloud registration (sPCR) would require extensive mapping of the environment. Also, localization based on camera image would face same issues as stereo ambiguities and hence depends on viewpoint. We thus introduce a novel approach to Joint Point Cloud and Image-based Localization (JPIL) for mixed-reality headsets that uses visual cues and headset orientation to register small, partially overlapped point clouds and save significant manual labor and time in environment mapping. Our empirical results compared to sPCR show average 10 fold reduction of required overlap surface area that could potentially save on average 20 minutes per inspection. JPIL is not only restricted to inspection tasks but also can be essential in enabling intuitive human-robot interaction for spatial mapping and scene understanding in conjunction with other agents like autonomous robotic systems that are increasingly being deployed in outdoor environments for applications like structural inspection.
	\end{abstract}

	%%%%%%%%%%%%%%%%%%%%%%%%%%%%%%%%%%%%%%%%%%%%%%%%%%%%%%%%%%%%%%%%%%%%%%%%%%%%%%%%
	\section{Introduction}
	The onset of portable mixed reality headsets like Google Glass and Microsoft HoloLens has enabled efficient human interactions with 3D information. These headsets are found to be suitable for on-site inspection~\cite{webster1996augmented,moreu2017augmented}, due to their ability to visualize augmented geo-tagged holograms and measure 3D distances by virtue of gaze and gesture. Notably, before every inspection, the headset needs to be localized, preferably with onboard sensors, to establish a common frame of reference for 3D coordinates. As spatial mapping and understanding is key to mixed-reality, we can safely assume that the primary sensors would include at least a depth sensor and a camera. However, due to low form factor and portability, these sensors have limited range. Given a prior 3D model of the structure as a template, existing methods for simple point cloud registration (sPCR)~\cite{pomerleau2015review,lei2017fast} can be employed on a spatial map generated on-site by the headset. These methods, however, require significant overlap between the point clouds and thus would require the user to sufficiently map the structure on the order of $300m^2$ surface area. Conversely, camera pose estimation using 3D-2D correspondences~\cite{6126302} lacks desired precision and depends on viewpoint due to stereo ambiguities.
	
	\begin{figure}[t]
		\centering
		\includegraphics[width=0.85\columnwidth]{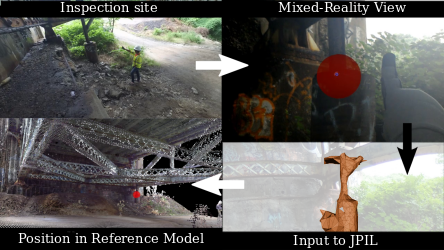}
		\caption{\small An inspector initiates JPIL with a gesture. JPIL uses spatial map, camera image and headset orientation for localization.}
		%  \vspace{-2.0em}
		\label{fig:Teaser}
		\vspace{-0.7cm}
	\end{figure}
	
	In this paper, we introduce an efficient approach to Joint Point Cloud and Image-based Localization (JPIL) for marker-free self-localization of mixed-reality headsets that requires minimum on-site mapping time. JPIL can successfully register spatial map with significantly low overlap with a prior 3D model by combining additional information from camera image and headset orientation, which is simply available from the onboard camera, IMU and magnetometer sensors. Interestingly, it simplifies the problem to an extent that is critical for operation in symmetric environments. A small spatial map might resemble multiple structures on the prior 3D model, thus, a camera pose estimation (CPE) problem is solved, as a function of point cloud registration to differentiate between multiple candidates. The resulting localization accuracy is similar, albeit at significantly lower point cloud overlap. The contributions of this paper are: 
	\begin{enumerate*}
		\item We contribute a JPIL method (Section~\ref{jpil}) to use spatial map (Section~\ref{pcr}), camera image (Section~\ref{imref}) and headset orientation to localize mixed-reality headset with minimum mapping time.
		\item A modified binary shape context based 3D point cloud descriptor with an efficient multiple-candidate descriptor matching algorithm (Section~\ref{matching}).
		\item We empirically show that JPIL method results in the successful registration of point clouds with overlap as low as $10m^2$ and average 10 fold reduction in required surface area.
	\end{enumerate*}
	
	Finally, we provide software implementation~\footnote{Source code: \url{https://bitbucket.org/castacks/jpil}} and hardware verification on a Microsoft HoloLens.
	
	\section{Digital Model and Setup}
	\label{digi_models}
\begin{figure}[h]
		\centering		
		\subfloat{\includegraphics[width=\columnwidth]{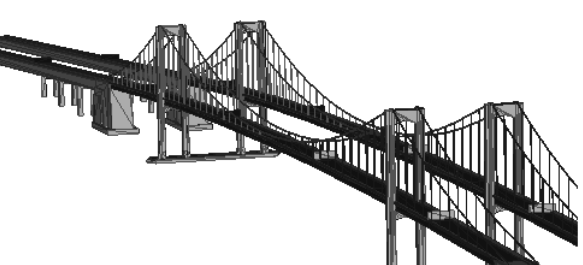}}
		\caption{\small An example CAD model $\mathcal{M}$ of the Delaware Memorial Bridge.}
		\label{fig:img-M}
	\end{figure}

	A common frame of reference is required to report 3D coordinates that can be persistently used across multiple inspection sessions. While a global positioning system (\textit{e.g.}, GPS) can be used, instead, we use a digital 3D model (\autoref{fig:img-M}) $\mathcal{M}$ of the structure and report all 3D coordinates in its reference frame $\mathcal{R}$. $\mathcal{M}$ has to be real scale and can be a partial or full representation of the structure. The source for $\mathcal{M}$ can be a computer-aided model or a map generated by other 3D reconstruction methods such as~\cite{geiger2011stereoscan, brenner2005building}. If $\mathcal{M}$ of a structure does not exist prior to its inspection, then the user can scan the structure to build a map that can be used as $\mathcal{M}$ in future inspections.

	We choose a 3D triangular mesh as the format for $\mathcal{M}$ as ray casting on a mesh for measurement is accurate and computationally cheap due to the surface representation. In most cases, a mixed-reality headset would setup up a local frame of reference $\mathcal{R}'$ and an arbitrarily defined origin $x'_0$ for its operation. Let $\mathcal{M'}$ define the triangle mesh generated by spatial mapping. Orientation $q$ is estimated in Easting, Northing and Elevation (ENU) geographic Cartesian frame from the IMU and magnetometer for both the models. The models are thus aligned with the ENU frame.
	
	Localization of the headset has to be performed only once per inspection session, where a transformation is established between $\mathcal{R}$ and $\mathcal{R}'$. Headset pose $x'_t$ w.r.t $\mathcal{R}'$ at time $t$ is tracked by the headset itself. A normal inspection procedure under the proposed method would be to first map a small ($\sim 5m^2$) structural surface $\mathcal{M'}$ and initiate JPIL with $(\mathcal{M},\mathcal{M'},x'_t,q_t)$, where $t$ represents the time in that instance.
	
	\section{Point Cloud Registration}
	\label{pcr}
\begin{figure}[h]
		\centering
		%\subfloat{\def\svgwidth{0.75\columnwidth}{\input{images/meshtransforms.pdf_tex}}}
		\subfloat{\def\svgwidth{0.75\columnwidth}{{\renewcommand\normalsize{\tiny}%
			\normalsize
			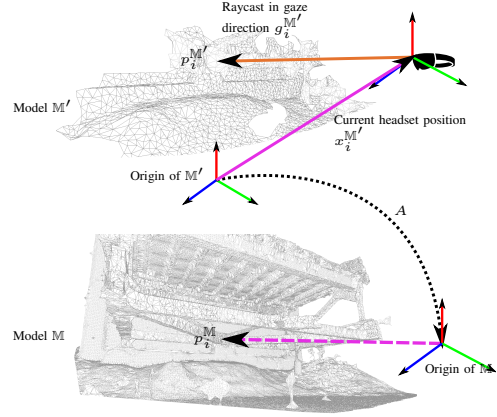} }}
		\caption{\small $\mathcal{R}$ establishes a common frame of reference at the origin of $\mathcal{M}$. During each inspection $\mathcal{M'}$ is generated, while $\mathcal{M}$ is common. A raycast performed in gaze direction $g_t^\mathcal{M'}$ from headset's position $x_t^\mathcal{M'}$ gives the measurement $p_t^\mathcal{M'}$. Point cloud registration estimates $A$, that allows the point of interest $p_t^\mathcal{M}$ to be reported in $\mathcal{R}$.}
		\label{fig:meshtransforms}
		\vspace{-0.2cm}
	\end{figure}

	JPIL samples $\mathcal{M}$ and $\mathcal{M'}$ meshes to generate corresponding point clouds for registration. Point cloud registration estimates a $4 \times 4$ transformation matrix $A$ given $\mathcal{M}$ and $\mathcal{M'}$, such that $\mathcal{M'}$ perfectly aligns with $\mathcal{M}$ when every point of $\mathcal{M'}$ is transformed by $A$~\autoref{fig:meshtransforms}.
	\begin{equation}
	p^{\mathcal{M}}_i = A p^{\mathcal{M'}}_i
	\end{equation}
	where $p^{\mathcal{M}}_i$ and $p^{\mathcal{M'}}_i$ are corresponding points (homogeneous coordinates) in $\mathcal{M}$ and $\mathcal{M'}$ respectively. A common framework consists of essentially the following steps: \begin{enumerate*}
		\item Keypoint extraction,
		\item Descriptor computation,
		\item Descriptor matching, and
		\item Transformation estimation.
	\end{enumerate*} The models are aligned with ENU frame, thus, $A$ will have negligible rotation component. We use binary shape context descriptor~\cite{bsc} for step 1 and 2 while we modify step 1 to incorporate the orientation information. The modified descriptor (tBSC) is now orientation specific. Finally, we propose an efficient algorithm for step 3 (\autoref{matching}). Thus, we discuss steps 1 and 3 in detail while briefly revising the other steps.
	
	\subsection{Keypoint extraction}
	\label{kp_detection}
	Keypoint extraction complements the feature descriptors by providing points that are expected to have high feature descriptiveness and are robust to minor viewpoint changes. For each point $p'$ of a point cloud, we perform two separate eigenvalue decompositions on the covariance matrices $M_{xy}$ and $M_{xyz}$.
	
	The covariance matrices $M_{xy}$ and $M_{xyz}$ are constructed using $p'$ and its neighboring points within a supporting radius $r^\textnormal{(scale)}$, which is user defined. We used $r^\textnormal{(scale)} = 0.4$. The neighboring points $q_j(j = 1,2,\ldots ,m)$ and $p'$ are denoted as $N = \{q_0,q_1,q_2,\ldots ,q_m\}$, where $q_0$ is the point $p'$ and $m$ is the number of neighboring points. A new set of points $N' = \{q'_0,q'_1,q'_2,\ldots,q'_m\}$ is generated from $N$ with only the $x$ and $y$ component of those points.
	The covariance matrices $M_{xyz}$ and $M_{xy}$ are calculated as follows:
	\begin{equation}
	\label{mxyz-def}
	M_{xyz} = \frac{1}{m+1}(q_j - q_0)(q_j - q_0)^\mathrm{T}
	\end{equation}
	\begin{equation}
	\label{mxy-def}
	M_{xy} = \frac{1}{m+1}(q'_j - q'_0)(q'_j - q'_0)^\mathrm{T}
	\end{equation}
	
	The eigenvalues in decreasing order of magnitude $\{\lambda_1, \lambda_2, \lambda_3\}$ and $\{\lambda'_1, \lambda'_2\}$, along with the corresponding eigenvectors $\{e_1, e_2, e_3\}$ and $\{e'_1, e'_2\}$ respectively, are computed by performing eigenvalue decompositions of $M_{xyz}$ and $M_{xy}$ respectively.
	
	Now, the point $p'$ is considered a keypoint if:
	\begin{itemize}
		\item $(\lambda'_2/\lambda'_1) < k$ : Note, $\{\lambda'_1, \lambda'_2\}$ are the eigenvalues computed from $M_{xy}$, which in turn is formed from the 2D distribution of $N'$ in the $x,y$ plane. This condition, thus, promotes points with well defined eigenvectors.
		\item Highest curvature among its neighbors. The curvature $c$ is defined by $c = \left(\frac{\lambda_3}{\lambda_1 + \lambda_2 + \lambda_3}\right)$.
	\end{itemize}
	
	For each keypoint, a local reference frame (LRF) is defined. An LRF imparts invariance in feature descriptors to rigid transformations. Descriptors are computed based on the LRFs, thus, it is critical for LRFs to be reproducible. In~\cite{bsc}, the authors define an LRF by adopting $p$ as the origin and $e_1, e_2, e_1 \times e_2$ as the $x\textnormal{-},y\textnormal{-}, \textnormal{ and } z\textnormal{-}\textnormal{axes}$, respectively, where $\times$ represents the cross product of vectors. However, the orientation of each eigenvector calculated by eigenvalue decomposition has a $180^\circ$ ambiguity. Therefore, there are two choices for the orientation of each axis and thus four possibilities for the LRF. They end up computing descriptors for all four possible LRFs and thus leading to reduced precision. This ambiguity is generally faced by all sPCR methods that use eigenvectors for LRF. In, contrast JPIL adopts $\hat{E}, \hat{N}, \hat{E} \times \hat{N}$ as the $x\textnormal{-},y\textnormal{-}, \textnormal{ and } z\textnormal{-}\textnormal{axes}$, respectively, where $\hat{E}$ and $\hat{N}$ is a vector towards magnetic East and North respectively. Thus, the LRF is written as
	\begin{equation}
	\{p, \hat{E}, \hat{N}, \hat{E} \times \hat{N}\}
	\end{equation}

	\subsection{Multiple Candidate Registration}
	\label{matching}
\begin{figure}[h]
 		\centering
 		\subfloat{\includegraphics[width=\columnwidth]{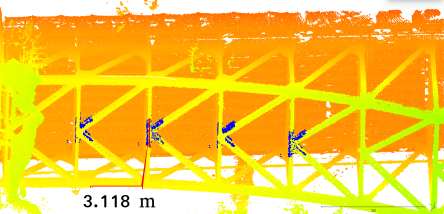}} 		
 		\caption{\small Multiple possible registration candidates for very small ($3m \times 1.5m \times 0.5m$) $\mathcal{M'}$ (Blue). A user can easily map such small region which is sufficient for unique localization when coupled with image based candidate selection.} 		
 		\label{fig:img-mcr} 		
 	\end{figure}

	 Since, $\mathcal{M'}$ would generally be very small as compared to $\mathcal{M}$, the registration would have translational symmetry~\autoref{fig:img-mcr}. Let the descriptor for a keypoint $p_i$ be defined as $d_i$ and $P_\mathcal{M} = \{ p_1^\mathcal{M}, p_2^\mathcal{M}, p_3^\mathcal{M}, \ldots, p_m^\mathcal{M} \}$ and $P_\mathcal{M'} = \{ p_1^\mathcal{M'}, p_2^\mathcal{M'}, p_3^\mathcal{M'}, \ldots, p_n^\mathcal{M'} \}$ be the set of keypoints for $\mathcal{M}$ and $\mathcal{M'}$ respectively.
	 
	 Let $C_{\textnormal{desc}}(d_i, d_j)$ be the hamming distance between two binary descriptors $d_i$ and $d_j$, and $c_{\textnormal{max}}$ denote maximum possible value of $C_{\textnormal{desc}}(d_i, d_j)$. A match $M$ is a set of keypoint pairs $\{p^\mathcal{M'}, p^\mathcal{M}\}$ formed by selecting a keypoint $p_i^\mathcal{M'}$ and $p_j^\mathcal{M}$ from $P_\mathcal{M'}$ and $P_\mathcal{M}$ respectively such that $C_{\textnormal{desc}}(d_i^\mathcal{M'}, d_j^\mathcal{M}) < c_{\textnormal{max}} \times \epsilon_{\textnormal{desc}}$, where $\epsilon_{\textnormal{desc}}$ is a user definable threshold. $M$ contains set of matches that 1) belong to either one of the possible candidates, and 2) are outliers. We find a family of match sets $O$ where a set represents a candidate. $O$ is found based on geometric consistency with error margin of $\epsilon_{\textnormal{clust}}$ and clustering on $M$ given by Algorithm~\ref{alg:CandidateClustering}. We used $\epsilon_{\textnormal{clust}} = 0.8$.
	 
	 \begin{algorithm}[]
	 	\caption{Find multiple candidate registrations}\label{alg:CandidateClustering}
	 	\begin{algorithmic}[1]
	 		\Procedure{FindRegCandidates}{$M, \epsilon_{\textnormal{clust}}$}
	 		\State $O\gets \{\}$ \Comment{Family of match sets}
	 		\State $A_\textnormal{all}\gets \{\}$ \Comment{Set of transformations}
	 		\State $C\gets \{\}$ \Comment{Set of alignment costs}
	 		\State $nos\gets|M|$	 		
	 		\State $gr\gets$ zeros$(1,nos)$
	 		\State $i\gets0$
	 		\While{$i<nos$}
	 		\If {$gr[i] == 0$}
	 			\State $\{p^\mathcal{M'},p^\mathcal{M}\}\gets$ $M[i]$
	 			\State $gr[i]\gets 1$
	 			\State $o\gets \{M[i]\}$
	 			\State $j\gets i+1$
	 			\While{$j<nos$}
	 			\If {$gr[j] == 0$}
	 				\State $\{q^\mathcal{M'},q^\mathcal{M}\}\gets$ $M[j]$
	 				\State $e_x \gets \abs{ p_x^\mathcal{M}-p_x^\mathcal{M'}-q_x^\mathcal{M}+q_x^\mathcal{M'} } $
	 				\State $e_y \gets \abs{ p_y^\mathcal{M}-p_y^\mathcal{M'}-q_y^\mathcal{M}+q_y^\mathcal{M'} } $
	 				\State $e_z \gets \abs{ p_z^\mathcal{M}-p_z^\mathcal{M'}-q_z^\mathcal{M}+q_z^\mathcal{M'} } $
	 				\If{$ e_x + e_y + ez < 3 \times \epsilon_{\textnormal{clust}}$}
 					  \State $o\gets o \cup \{M[j]\}$
 					  \State $gr[j] \gets 1$
	 				\EndIf	 				
	 			\EndIf
	 			\EndWhile
	 			\If {$|o| > 4$}
	 				\State $a\gets$ \Call{RigidTransform}{$x$}
	 				\State $c\gets$ \Call{AlignCost}{$a$}
	 				\State $O\gets O \cup \{o\}$
	 				\State $C\gets C \cup \{c\}$
	 				\State $A_\textnormal{all}\gets A_\textnormal{all} \cup \{a\}$ 				
	 			\EndIf
	 		\EndIf
	 		\State $i\gets i+1$
	 		\EndWhile
	 		\State \Return $O, A, C$
	 		\EndProcedure
	 	\end{algorithmic}
	 \end{algorithm}
 	
 	Subroutine \(\Call{RigidTransform}{o}\) called by Algorithm~\ref{alg:CandidateClustering} estimates the transformation matrix $a$ from the corresponding 3D points using singular value decomposition. An alignment is evaluated based on how similar the two point clouds looks after alignment. $\mathcal{M'}$ is transformed by $a$ to get $\mathcal{M'}_a$, and $\mathcal{M}$ is clipped to get $\mathcal{M}_c$ using a box filter of dimension equivalent to the bounding box of $\mathcal{M'}_a$. The idea is to compute two tBSC descriptors $d_1$ and $d_2$, one each for $\mathcal{M'}_a$ and $\mathcal{M}_c$ with radius equivalent to longest side of the bounding box, and keypoints $p_1$ and $p_2$ at the box centers respectively. Subroutine \(\Call{AlignCost}{a}\) return the alignment cost given by $C_{\textnormal{desc}}(d_1, d_2)$.
 	
	\section{Image Based Candidate Selection}
	\label{imref}
	The motivation behind JPIL to use visual cues is the rich amount of information that is present in a single image. While the depth sensor on the headsets has a range of about $5m$ and about $70^\circ$ field of view (FOV), a spherical camera, can provide full $360^\circ$ horizontal and $180^\circ$ vertical field of view. Consider a particular candidate $j$: let its transformation matrix and alignment cost be $a_j$ and $c_j$ respectively. Since, the headset pose $x'_t$ w.r.t $\mathcal{M'}$ at time $t$ is known (Section~\ref{digi_models}). JPIL generates synthetic projection images $\mathcal{I}_{t,j}^{\mathcal{M}}(q_t,x_{t,j})$ of $\mathcal{M}$ setting a virtual camera at position $x_{t,j} = a_j \times x'_t$ and headset orientation $q_t = \{q_t^\textnormal{Roll}, q_t^\textnormal{Pitch}, q_t^\textnormal{Yaw}\}$ in ENU frame.
	
	If $j$ is the correct candidate, then the camera image $\mathcal{I}_{t}$ and synthetic image $\mathcal{I}_{t,j}^{\mathcal{M}}$ should match better than those of the other candidates. Thus, implying $x_{t,j}$ to be the correct headset pose w.r.t $\mathcal{R}$. We evaluate a match based on the distance metric described in Section~\ref{3d2d}.
	
	We demonstrate JPIL with a spherical camera, however it is not a necessity. The user may use any camera with suitable FOV for their case. We however do recommend the use of spherical cameras as the localizability increases with FOV and we discuss the benefits of the same in the experiments.
	
	\subsection{3D-2D Image Match Distance Metric}
	\label{3d2d}
		Given a 3D point cloud structure and a camera image, the estimation of the camera pose is a standard computer vision problem. We follow interesting articles~\cite{6126302,activ} from the literature that solves this problem. And use it as a framework to build upon our method that supports spherical image projection and a orientation constraint non-linear optimization for camera pose.
		
\begin{figure}[h]
			\centering
			\subfloat{\def\svgwidth{0.7\columnwidth}{{\renewcommand\normalsize{\small}%
						\normalsize
						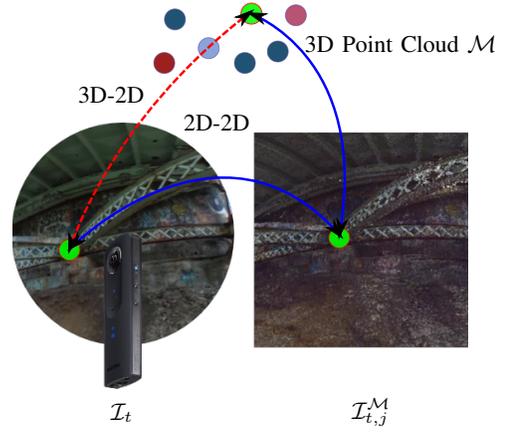} }}
			\caption{\small Generating 3D-2D correspondence from 2D-2D image correspondences and backtracking to point cloud.}
			\label{fig:im_correspondences}
		\end{figure}

		Since, $\mathcal{I}_{t,j}^{\mathcal{M}}$ is generated by projecting $\mathcal{M}$ on a sphere, we can backtrack from a pixel coordinate $(x,y)$ to the 3D points $p_{x,y} = f(\mathcal{I}_{t,j}^{\mathcal{M}},x,y)$. The initial goal is to detect 2D-2D image correspondences between $\mathcal{I}_{t,j}^{\mathcal{M}}$ and $\mathcal{I}_{t}$, and establish 3D-2D correspondences after backtracking from $\mathcal{I}_{t,j}^{\mathcal{M}}$~\autoref{fig:im_correspondences}. 
		
		Let $\mathbf{P}_{x,y} = \{p | p = p_{x+i_x,y+i_y}, i_x \in [-n,n], i_y \in [-n,n] \}$ where $n$ is kernel dimension. Standard deviation $\sigma_{x,y} = std(\mathbf{P}_{x,y})$. JPIL rejects a correspondence if $\sigma_{x,y} < \epsilon_\sigma$, assuming the 3D point to be uncertain.
		
		Let $p_k$ and $\bar{p_k}$ be 3D point and image pixel respectively of $k^{th}$ correspondence. $x'_{t,j}$ is the position of headset in $\mathcal{M}$ according to $j^{th}$ candidate. Using spherical projection, we can project $\bar{p_k}$ to a point $\bar{p_k^\textnormal{s}}$ on the projection sphere. Also, we can project every point $p_k$ to a point $p_k^\textnormal{s}(q_o,x'_o)$ on projection sphere as a function of orientation $q_o = \{q_o^\textnormal{Roll}, q_o^\textnormal{Pitch}, q_o^\textnormal{Yaw}\}$ and position $x_o$ (viewpoint). Thus, we can solve for $\{q_o,x_o\}$ by minimizing the cost $C_{\textnormal{SnP}(q_o,x_o)}$ on the set of 3D-2D correspondences with RANSAC~\cite{fischler1987random} based outlier rejection.
		\begin{equation}
			\label{optimization_eqn}
			C_{\textnormal{SnP}} = \frac{1}{2}\sum_k\left( \norm{p_k^s(q_o,x_o) - \bar{p_k^s}}^2 \right)
		\end{equation}
		\begin{equation*}
		    q_t^\alpha-\epsilon_q < q_o^\alpha < q_t^\alpha+\epsilon_q
		\end{equation*}
		where $\alpha$ is Roll, Pitch or Yaw angle and $\epsilon_q$ is the allowed flexibility. We use Ceres solver~\cite{agarwal2012ceres} for the optimization. To evaluate the similarity of two images there are many distance metrics in the literature, like Hausdorff distance~\cite{zhao2005new}, photometric error~\cite{torreao1998matching} and homography consistent percentage inlier matches. Since, we have the 3D information of an image, we rather check for geometric consistency with the error in position given by $C_{\textnormal{image}}$:
		\begin{equation}
			C_{\textnormal{image}}(\mathcal{I}_{t}, \mathcal{I}_{t,j}^{\mathcal{M}}) = \norm {x_{t,j} - x_o}
		\end{equation}
		
	\subsection{Confident Positive Matches}
		The cost function $C_{\textnormal{SnP}}$ is non-linear and due to noise in feature matching, the optimization might reach a false local minimum giving erroneous $x_o$ estimate. Therefore, $C_{\textnormal{image}}$ metric is only good to determine if a match is confident positive $\mathcal{L}_+$ given by:
		\begin{equation}
			\mathcal{L}_+(\mathcal{I}_{t}, \mathcal{I}_{t,j}^{\mathcal{M}}) = \begin{cases}
				1, & \text{if }C_{\textnormal{image}}(\mathcal{I}_{t}, \mathcal{I}_{t,j}^{\mathcal{M}}) \le \epsilon^+\\
				0, & \text{otherwise}
				\end{cases}
		\end{equation}
		where $\epsilon^+$ depends on the noise level and we used $\epsilon^+ = 1.5$.
	
	\section{Joint Point Cloud and Image Based Localization}
	\label{jpil}
	To summarize, JPIL has as its inputs, a reference model $\mathcal{M}$, a small 3D map of the structure scanned in the particular session $\mathcal{M'}$, headset position $x'_t$, headset orientation in ENU frame $q_t$ and spherical image $\mathcal{I}_{t}$ at time $t$. The output of JPIL Algorithm~\ref{alg:JPIL} is $A$, the transformation of $\mathcal{M'}$ to $\mathcal{M}$, such that headset position $x_t = A \times x'_t$ w.r.t $\mathcal{R}$ can be estimated.
	
	\begin{algorithm}[]
		\caption{Localize Headset w.r.t $\mathcal{R}$}\label{alg:JPIL}
		\begin{algorithmic}[1]
			\Procedure{JPIL}{$\mathcal{M}, \mathcal{M'}, x'_t, q_t, \mathcal{I}_{t}$}
				\State $\alpha^t \gets$ registration error threshold 
				\State $P_\mathcal{M}\gets$ set of keypoints from $\mathcal{M}$ 
				\State $P_\mathcal{M'}\gets$ set of keypoints from $\mathcal{M'}$ 
				\State $M\gets $ Descriptor match of $P_\mathcal{M}$ and $P_\mathcal{M'}$
				\State $[O, A_\textnormal{all}, C] \gets \Call{FindRegCandidates}{M, \alpha^t}$
				\State $n \gets |A_\textnormal{all}|$
				\State $j \gets 0$
				\While {$j < n$}
					\State $\mathcal{I}_{t,j}^{\mathcal{M}} \gets$ Synthetic spherical image for $A_\textnormal{all}[j]$
					\If{$\mathcal{L}_+(\mathcal{I}_{t}, \mathcal{I}_{t,j}^{\mathcal{M}})$}
						\State \Return $A_\textnormal{all}[j]$
					\EndIf
				\EndWhile
				\State $i \gets \argmin(C)$
				\State \Return $A_\textnormal{all}[i]$
			\EndProcedure
		\end{algorithmic}
	\end{algorithm}

	\section{Experimental Results}
	\label{experiments}
	We performed few experiments to evaluate the following:
	\begin{enumerate}
		\item Performance of CPE with orientation constraints and differentiability between candidates.
		\item Tolerance of tBSC to error in orientation.
		\item Relation of $C_{\textnormal{image}}(\mathcal{I}_{t}, \mathcal{I}_{t,j}^{\mathcal{M}})$ to error in orientation and relative distance between the camera poses of the two images.
		\item Relation of $C_{\textnormal{image}}(\mathcal{I}_{t}, \mathcal{I}_{t,j}^{\mathcal{M}})$ to $\epsilon^\sigma$.
		\item Reduction in minimum required surface area and mapping time.
	\end{enumerate}
	
\begin{figure}[h]
		\vspace{-0.4cm}	
		\centering
		\subfloat{\def\svgwidth{0.43\columnwidth}{{\renewcommand\normalsize{\small}%
					\normalsize
					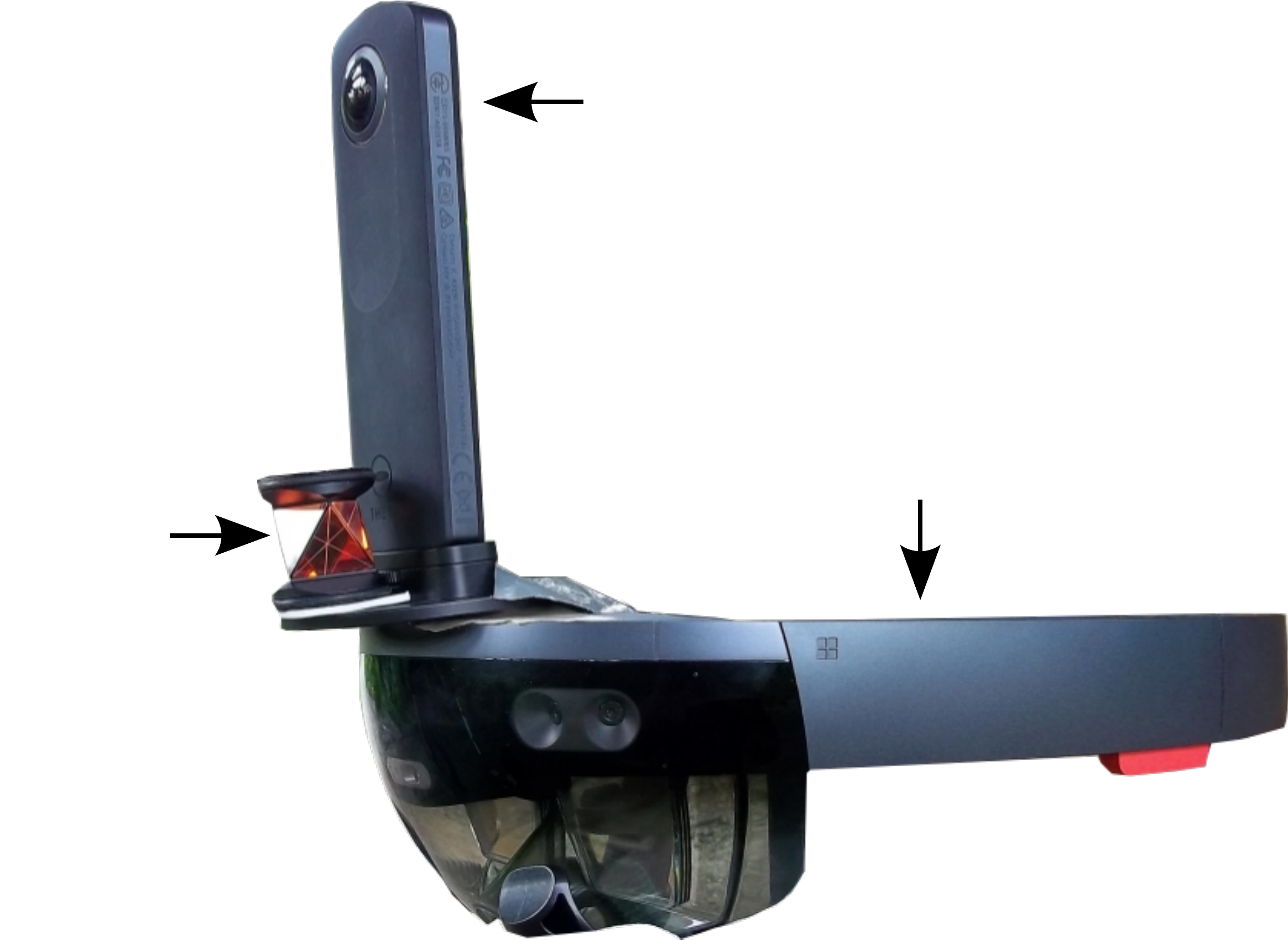} }}
		\subfloat{\includegraphics[width=0.5\columnwidth]{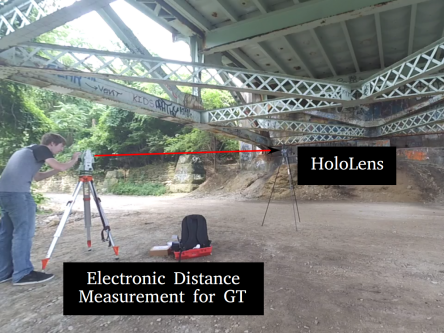}}
		\caption{\small Microsoft HoloLens used for the experiments and test arena.}
		\label{fig:img-hololens}

	\end{figure}

	The experiments were performed on a Microsoft HoloLens headset with a Ricoh theta S $360^\circ$ camera~\autoref{fig:img-hololens}. We used an off-board computer to process the data. The HoloLens uploaded $\mathcal{M'}$ ($\sim10$ MB), $\mathcal{I}_{t}$ ($\sim700$ kB for $1280 \times 640$ pixels), $x'_t$ and $q_t$ to the server. $\mathcal{M}$ ($\sim1.2$ GB) was already present in the server and an instance of JPIL was running to accept communication from the HoloLens and return back the transformation $A$. We used SPHORB~\cite{zhao2015sphorb} feature descriptors for the spherical images.
	
	We tested JPIL in real world as well as simulated environments. Real world data from HoloLens was collected from Charles Anderson bridge, Pittsburgh, PA, USA. We used a high precision Terrestrial Laser Scanner (TLS) to generate dense RGB $\mathcal{M}$. The ground truth positions for the experiments were measured using a Electronic Distance Measuring (EDM) Survey equipment that can report the position of an EDM prism with millimeter accuracy. We manually calibrated the EDM measurement w.r.t $\mathcal{R}$.
	
	\subsection{Performance of CPE with orientation constraints and differentiability between candidates.}
\begin{figure}[h]
			\centering
			\vspace{-0.3cm}
			\subfloat[]{\includegraphics[width=0.5\columnwidth]{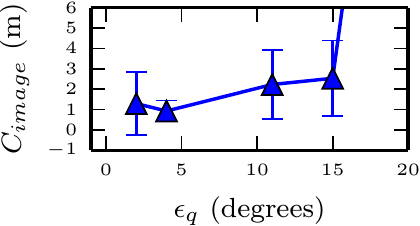}}
			\subfloat[]{\includegraphics[width=0.5\columnwidth]{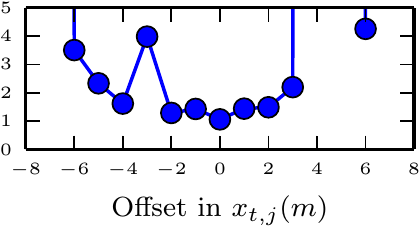}}
			\vspace{-0.1cm}
			\caption{\small CPE error with varying relaxation of orientation constraints (a) and error in nominal position (b).}
			\label{fig:imrefVeqd}
			\vspace{-0.2cm}
		\end{figure}

		The cost function~\eqref{optimization_eqn} might have many local minimums due to erroneous feature matching that might get selected by RANSAC as inliers. We performed CPE for 22 image pairs with varying $\epsilon_q$ and computed the average error and standard deviation for each $\epsilon_q$~\autoref{fig:imrefVeqd}. We observed that errors increase drastically with $\epsilon_q > 15^\circ$ as expected. We observed a slight decrease at $\epsilon_q = 4^\circ$, which can be credited to the flexibility that allowed optimization to minimize considering errors in $q_t$ estimate. We also evaluate how discriminative is CPE to candidate positions. We took a spherical image $\mathcal{I}_{t}$ and generated multiple spherical images $\mathcal{I}_{t}^{\mathcal{M}}$ at position with error increments of $\pm0.5m$ along the bridge. From~\autoref{fig:imrefVeqd} we observe that $C_{\textnormal{image}}(\mathcal{I}_{t}, \mathcal{I}_{t,j}^{\mathcal{M}})$ becomes unstable with increasing error and thus the concept of confident positive matching works well to discriminate between candidates that are further away from nominal position.
				
	\subsection{Tolerance of tBSC to error in orientation.}
\begin{figure}[h]
		\centering
		\vspace{-0.1cm}
		\subfloat{\includegraphics[width=0.9\columnwidth]{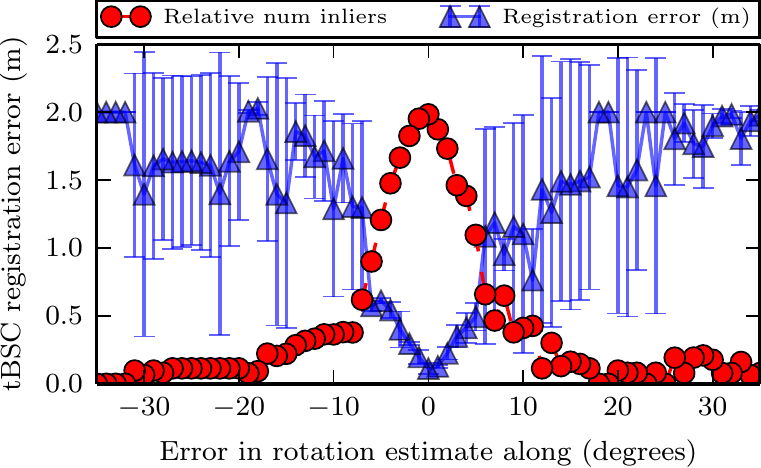}}
		\caption{\small tBSC registration error with error in orientation estimates.}
		\label{fig:tBSCvO}
		\vspace{-0.2cm}
	\end{figure}

	We added error in the orientation estimates along X, Y and Z axes individually in an increment of $\pm1^\circ$. We, then performed tBSC registration and selected the candidate with transformation estimate closest to the ground truth transformation. The results in~\autoref{fig:tBSCvO} shows a minimum tolerance of $\pm5^\circ$ for error within 0.6m. It indicates the rotation specificity of LRF as well as robustness to error in sensor values.
	
	\subsection{Relation of $C_{\textnormal{image}}(\mathcal{I}_{t}, \mathcal{I}_{t,j}^{\mathcal{M}})$ to error in orientation.}
\begin{figure}[h]
		\centering
		%\vspace{-0.3cm}
		\subfloat{\includegraphics[width=0.85\columnwidth]{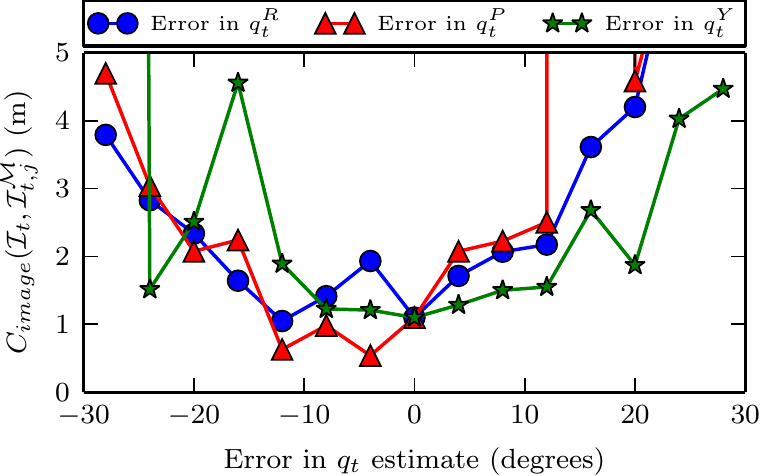}}
		\caption{\small Performance in CPE with error in orientation estimate.}
		\label{fig:imrefVq}
		\vspace{-0.1cm}
	\end{figure}

	$\mathcal{I}_{t,j}^{\mathcal{M}}(q_t,x_{t,j})$ is generated as a function of $(q_t,x_{t,j})$. We wanted to evaluate how error in CPE $C_{\textnormal{image}}(\mathcal{I}_{t}, \mathcal{I}_{t,j}^{\mathcal{M}})$ is affected by error in $q_t$. So, we added errors in $q_t=q_t+q_e$ and performed CPE for each pair of $\mathcal{I}_{t}$ and $\mathcal{I}_{t,j}^{\mathcal{M}}(q_t+q_e,x_{t,j})$. We observe that CPE is tolerant to error in orientation estimates up to  $\pm10^\circ$~\autoref{fig:imrefVq}.	
	
	\subsection{Relation of $C_{\textnormal{image}}(\mathcal{I}_{t}, \mathcal{I}_{t,j}^{\mathcal{M}})$ to $\epsilon_\sigma$}
\begin{figure}[h]
		\centering		
		\subfloat{\includegraphics[width=0.75\columnwidth]{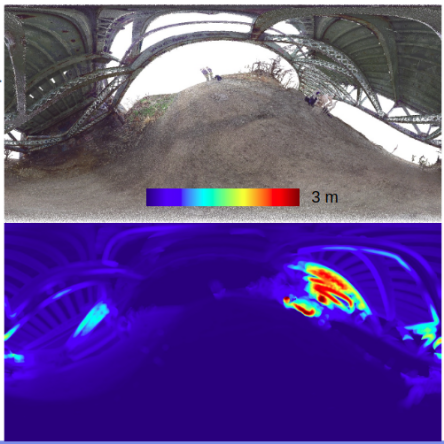}}
		\caption{\small Heatmap visualization of standard deviation (m) in $\mathcal{I}_{i,j}^{\mathcal{M}}$}
		\label{fig:heatmap}
		\vspace{-0.2cm}
	\end{figure}

\begin{figure}[h]
		\centering
		%\vspace{-0.4cm}
		\subfloat{\includegraphics[width=0.85\columnwidth]{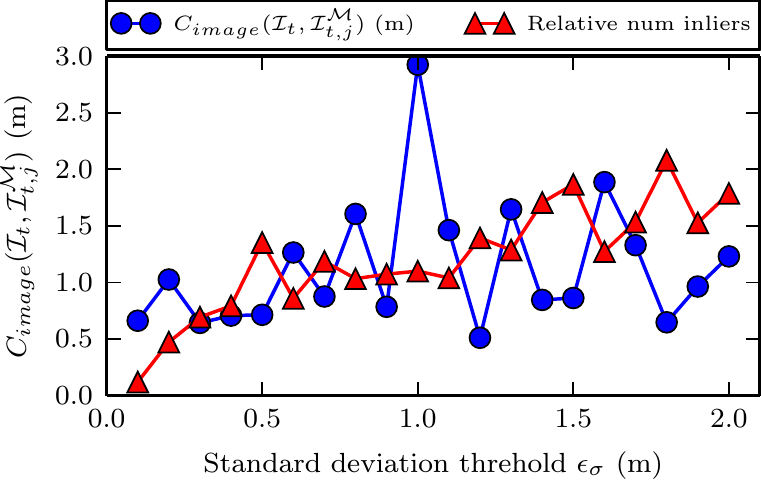}}
		\vspace{-0.1cm}
		\caption{\small With increasing $\epsilon_\sigma$, the error in CPE tends to increase, however, more inlier matches are being used which contributes to confident outlier rejection.}
		\label{fig:imrefVes}
		\vspace{-0.2cm}
	\end{figure}

	The~\autoref{fig:heatmap} shows an example synthetic spherical image and a heatmap visualization of $\sigma_{x,y}$ at each of its pixel. $\sigma_{x,y}$ gives the uncertainty measure of 3D-2D correspondence. Correspondences with high $\sigma_{x,y}$ value might result in more erroneous CPE, while a generous threshold would promote the number of inliers the can constrain the camera pose. We performed CPE with varying $\epsilon_\sigma$ on an image pair. From~\autoref{fig:imrefVes} we can observe that the variance of $C_{\textnormal{image}}(\mathcal{I}_{t}, \mathcal{I}_{t,j}^{\mathcal{M}})$ certainly increases, however relative number of inliers increase too.

	\subsection{Reduction in minimum required surface area and mapping time.}
\begin{figure}[h]
		\centering
		\vspace{-0.3cm}
		\subfloat{\includegraphics[width=0.5\columnwidth]{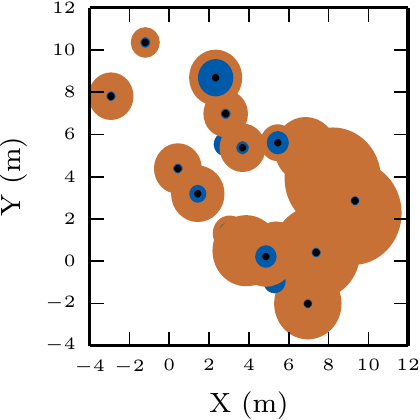}}
		\subfloat{\includegraphics[width=0.5\columnwidth]{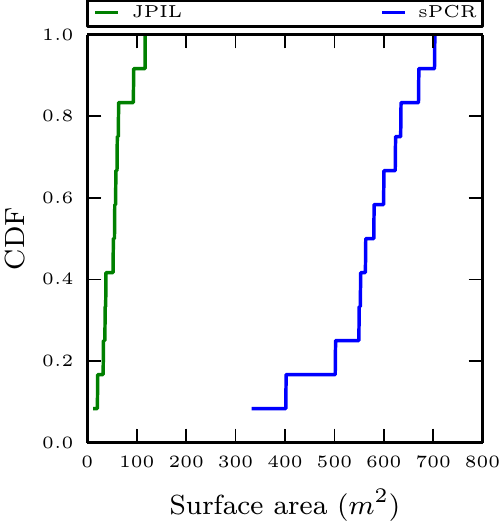}}		
		\caption{\small \textbf{Left:} Overall JPIL performance in X and Y axis. \textit{Black} is the ground truth, \textit{blue} is error in tBSC registration and \textit{brown} is error in CPE. \textbf{Right: } Surface area required for successful registration by JPIL and sPCR in 15 real world datasets.}
		\label{fig:cdf_walks}
		%\vspace{-0.4cm}
	\end{figure}

\begin{figure*}[t!]
		\centering
		\subfloat{\includegraphics[width=5.8cm]{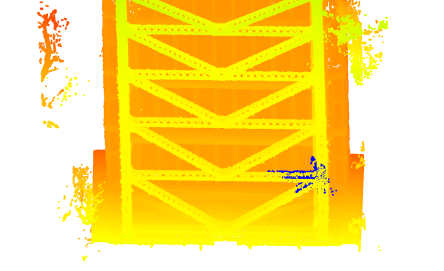}}
		\hspace{0.005cm}
		\subfloat{\includegraphics[width=5.8cm]{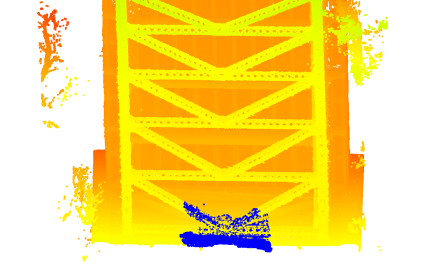}}
		\hspace{0.005cm}
		\subfloat{\includegraphics[width=5.8cm]{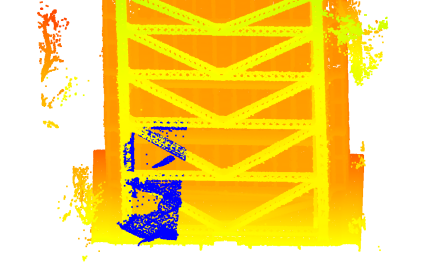}}
		\vspace{0.005cm}
		\addtocounter{subfigure}{-3}
		\subfloat[$\mathcal{S}=5m^2$]{\includegraphics[width=5.8cm]{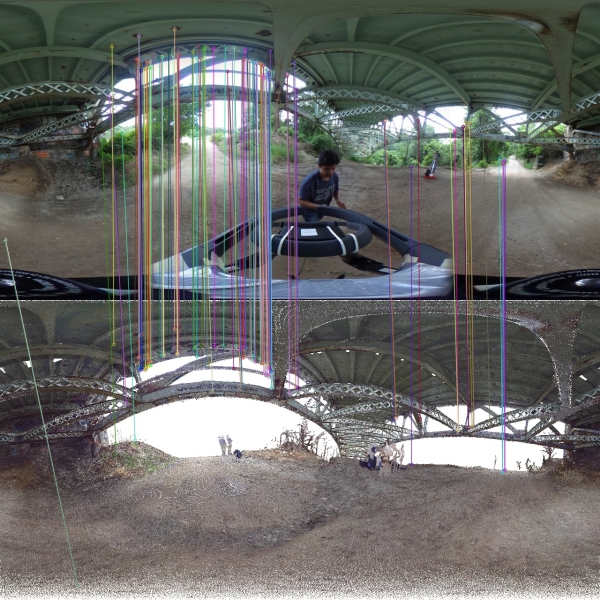}}
		\hspace{0.005cm}
		\subfloat[$\mathcal{S}=40m^2$]{\includegraphics[width=5.8cm]{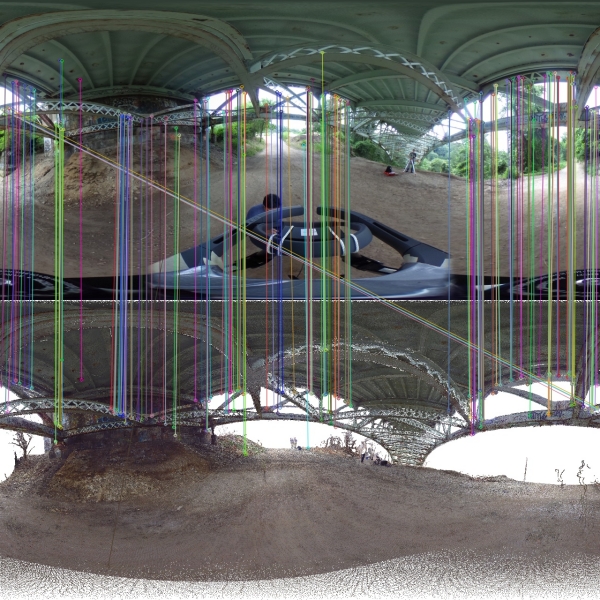}}
		\hspace{0.005cm}
		\subfloat[$\mathcal{S}=50m^2$]{\includegraphics[width=5.8cm]{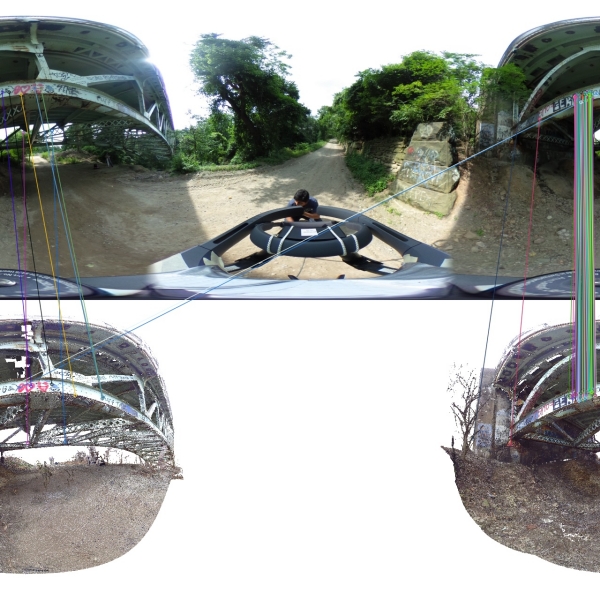}}
		
		\caption{\small\textbf{Top:} Example JPIL runs for three inspection sessions. \textit{Blue} denotes $\mathcal{M}'$, the spatial map generated by the HoloLens and \textit{yellow} denotes the reference model $\mathcal{M}$. Synthetic spherical images \textbf{Bottom} of confident positive matches are shown along with real image \textbf{Middle}. The three spatial maps shown here have varying surface area $\mathcal{S}$.}
		\label{fig:jpilexamples}
		\vspace{-0.4cm}
	\end{figure*}

	JPIL is targeted towards enabling accurate localization for very low overlap point clouds that requires significantly less user time. In sPCR, a user would need to walk on-site while mapping the structure to build a sufficiently large map. We simulated random walks in the vicinity of point clouds generated by the HoloLens. As the walk distance increased, more parts of the HoloLens point cloud were included for registration. We wanted to evaluate tBSC and BSC as a function of surface area mapped. Thus, we generated a cumulative density function of minimum surface area required by these methods for successful localization on 15 real datasets~\autoref{fig:cdf_walks}. We observe an average reduction of 10 times the surface area required by sPCR. For one dataset, we achieved a reduction from $465.50m^2$ to $13.226m^2$ and the time difference was $\sim20$ minutes. Finally, performing JPIL over 12 datasets with EDM ground truth, we observe an average accuracy of 0.28m for tBSC registration and 1.21m for CPE~\autoref{fig:jpilexamples}. The surface area was calculated by remeshing the point clouds and summing up the area of each triangle. The surface also included parts of the environment other than the structure, thus the required surface area in practice would be less than the values shown.
	%\vspace{-0.1cm}
	
	\section{Related Works}
	While we have covered the related works in the above text, here we emphasize on few other point cloud registration methods that uses visual cues. Dold~\cite{dold2006registration} uses planar patches from image data to refine a point cloud registration whereas Men \textit{et al}.~\cite{men2011color} uses hue as fourth dimension (x,y,z,hue) and search for correspondence in 4D space. Similarly, authors of~\cite{han2013lidar,al2006registration} use 2D image features in a tightly coupled framework to direct point cloud registration. These requires accurate calibration between the Lidar scans and camera images and work well for dominantly planer structures without stereo ambiguities. When accounted for errors in sensor calibration, stereo ambiguities and complex 3D environments, local image features tend to fail and thus decrease the robustness due to the their tightly coupled nature.
	
	\section{Conclusion And Future Work}
	We have presented a marker-free self-localization method for mixed-reality headsets and emphasized that data from three onboard sensors: a depth sensor, a camera and an IMU unit are critical for efficient localization. Our method is robust against errors in orientation estimation unto $\pm5^\circ$, which is generous for most sensors. Localization accuracy of 0.28m, is comparable to that of sPCR while requiring 10 fold less surface area on average. Our method does not evaluate the user's selection of $\mathcal{M}'$. Practically, the user should generate $\mathcal{M}'$ from well defined structures with minimum symmetry, and which also exists in $\mathcal{M}$. In future, we would like to explore the 3D information from time series image data to further enhance the efficiency and robustness of this method.
	
	\section{Acknowledgment}
	The authors would like to thank Daniel Maturana of Carnegie Mellon University for his inputs in the initial phase of the framework development and also proofreading.

	\bibliographystyle{ieeetr}
	\bibliography{ref}
	
	%\balance
	
	
\end{document}

%% file: images/meshtransforms.pdf_tex
%% Creator: Inkscape inkscape 0.48.4, www.inkscape.org
%% Accompanies image file 'meshtransforms.pdf' (pdf, eps, ps)
\begingroup%
  \makeatletter%
  \providecommand\color[2][]{%
    \errmessage{(Inkscape) Color is used for the text in Inkscape, but the package 'color.sty' is not loaded}%
    \renewcommand\color[2][]{}%
  }%
  \providecommand\transparent[1]{%
    \errmessage{(Inkscape) Transparency is used (non-zero) for the text in Inkscape, but the package 'transparent.sty' is not loaded}%
    \renewcommand\transparent[1]{}%
  }%
  \providecommand\rotatebox[2]{#2}%
  \ifx\svgwidth\undefined%
    \setlength{\unitlength}{1320.37880859bp}%
    \ifx\svgscale\undefined%
      \relax%
    \else%
      \setlength{\unitlength}{\unitlength * \real{\svgscale}}%
    \fi%
  \else%
    \setlength{\unitlength}{\svgwidth}%
  \fi%
  \global\let\svgwidth\undefined%
  \global\let\svgscale\undefined%
  \makeatother%
  \begin{picture}(1,0.87543266)%
    \put(0,0){\includegraphics[width=\unitlength]{meshtransforms.pdf}}%
    \put(0.00985381,0.62740971){\color[rgb]{0,0,0}\makebox(0,0)[lb]{\smash{Model $\mathbb{M'}$}}}%
    \put(0.00914854,0.15958836){\color[rgb]{0,0,0}\makebox(0,0)[lb]{\smash{Model $\mathbb{M}$}}}%
    \put(0.40254328,0.48690624){\color[rgb]{0,0,0}\makebox(0,0)[rb]{\smash{Origin of $\mathbb{M'}$}}}%
    \put(0.85469517,0.09136412){\color[rgb]{0,0,0}\makebox(0,0)[lb]{\smash{Origin of $\mathbb{M}$}}}%
    \put(0.41485653,0.72715672){\color[rgb]{0,0,0}\makebox(0,0)[rb]{\smash{$p_i^\mathbb{M'}$}}}%
    \put(0.42791335,0.15557754){\color[rgb]{0,0,0}\makebox(0,0)[rb]{\smash{$p_i^\mathbb{M}$}}}%
    \put(0.79082441,0.41477906){\color[rgb]{0,0,0}\makebox(0,0)[lb]{\smash{$A$}}}%
    \put(0.43851571,0.85282865){\color[rgb]{0,0,0}\makebox(0,0)[lt]{\begin{minipage}{0.27590671\unitlength}\raggedright Raycast in gaze  direction $g_i^\mathbb{M'}$ \end{minipage}}}%
    \put(0.66808935,0.62039459){\color[rgb]{0,0,0}\makebox(0,0)[lt]{\begin{minipage}{0.28104779\unitlength}\raggedright Current headset  position $x_i^\mathbb{M'}$\end{minipage}}}%
  \end{picture}%
\endgroup%

%% file: images/imref.pdf_tex
%% Creator: Inkscape inkscape 0.48.4, www.inkscape.org
%% Accompanies image file 'imref.pdf' (pdf, eps, ps)
\begingroup%
  \makeatletter%
  \providecommand\color[2][]{%
    \errmessage{(Inkscape) Color is used for the text in Inkscape, but the package 'color.sty' is not loaded}%
    \renewcommand\color[2][]{}%
  }%
  \providecommand\transparent[1]{%
    \errmessage{(Inkscape) Transparency is used (non-zero) for the text in Inkscape, but the package 'transparent.sty' is not loaded}%
    \renewcommand\transparent[1]{}%
  }%
  \providecommand\rotatebox[2]{#2}%
  \ifx\svgwidth\undefined%
    \setlength{\unitlength}{442.23999023bp}%
    \ifx\svgscale\undefined%
      \relax%
    \else%
      \setlength{\unitlength}{\unitlength * \real{\svgscale}}%
    \fi%
  \else%
    \setlength{\unitlength}{\svgwidth}%
  \fi%
  \global\let\svgwidth\undefined%
  \global\let\svgscale\undefined%
  \makeatother%
  \begin{picture}(1,0.92182763)%
    \put(0,0){\includegraphics[width=\unitlength]{imref.pdf}}%
    \put(0.64218526,0.81287698){\color[rgb]{0,0,0}\makebox(0,0)[lb]{\smash{3D Point Cloud $\mathcal{M}$}}}%
    \put(0.78830219,0.00607523){\color[rgb]{0,0,0}\makebox(0,0)[b]{\smash{$\mathcal{I}_{t,j}^{\mathcal{M}}$}}}%
    \put(0.24001392,0.00426626){\color[rgb]{0,0,0}\makebox(0,0)[b]{\smash{$\mathcal{I}_{t}$}}}%
    \put(0.44818709,0.64102459){\color[rgb]{0,0,0}\makebox(0,0)[b]{\smash{2D-2D}}}%
    \put(0.28872898,0.70094681){\color[rgb]{0,0,0}\makebox(0,0)[rb]{\smash{3D-2D}}}%
  \end{picture}%
\endgroup%

%% file: images/HLsetup.pdf_tex
%% Creator: Inkscape inkscape 0.48.4, www.inkscape.org
%% Accompanies image file 'HLsetup.pdf' (pdf, eps, ps)
\begingroup%
  \makeatletter%
  \providecommand\color[2][]{%
    \errmessage{(Inkscape) Color is used for the text in Inkscape, but the package 'color.sty' is not loaded}%
    \renewcommand\color[2][]{}%
  }%
  \providecommand\transparent[1]{%
    \errmessage{(Inkscape) Transparency is used (non-zero) for the text in Inkscape, but the package 'transparent.sty' is not loaded}%
    \renewcommand\transparent[1]{}%
  }%
  \providecommand\rotatebox[2]{#2}%
  \ifx\svgwidth\undefined%
    \setlength{\unitlength}{670.70537109bp}%
    \ifx\svgscale\undefined%
      \relax%
    \else%
      \setlength{\unitlength}{\unitlength * \real{\svgscale}}%
    \fi%
  \else%
    \setlength{\unitlength}{\svgwidth}%
  \fi%
  \global\let\svgwidth\undefined%
  \global\let\svgscale\undefined%
  \makeatother%
  \begin{picture}(1,0.72925593)%
    \put(0,0){\includegraphics[width=\unitlength]{HLsetup.pdf}}%
    \put(0.72245763,0.37014904){\color[rgb]{0,0,0}\makebox(0,0)[b]{\smash{HoloLens}}}%
    \put(-0.32994317,0.36480415){\color[rgb]{0,0,0}\makebox(0,0)[lt]{\begin{minipage}{0.43962247\unitlength}\raggedleft EDM \\ prism\end{minipage}}}%
    \put(0.46782834,0.68737868){\color[rgb]{0,0,0}\makebox(0,0)[lt]{\begin{minipage}{0.52482061\unitlength}\raggedright Theta $360^\circ$\\ camera\end{minipage}}}%
  \end{picture}%
\endgroup%